\documentclass[letterpaper, 10 pt, conference]{ieeeconf}  

\IEEEoverridecommandlockouts                              

\usepackage{amsmath}
\usepackage{booktabs}

\usepackage{amsmath}
\usepackage{amssymb}
\usepackage{amsfonts}
\usepackage{CJKutf8}
\usepackage{graphicx}
\usepackage{subcaption}
\usepackage{xcolor}
\usepackage[labelsep=period]{caption}

\title{\LARGE \bf
FreeDriveRF: Monocular RGB Dynamic NeRF without Poses for Autonomous Driving via Point-Level Dynamic-Static Decoupling
}

\author{Yue Wen$^{1}$, Liang Song$^{2}$, Yijia Liu$^{3}$, Siting Zhu$^{1}$, Yanzi Miao$^{3}$, Lijun Han$^{1}$, and Hesheng Wang$^{1}$
\thanks{This work was supported in part by the Natural Science Foundation of China under Grant 62225309, U24A20278, 62361166632, U21A20480. Corresponding Author: Hesheng Wang.}
\thanks{$^{1}$Department of Automation, Key Laboratory of System Control and
Information Processing of Ministry of Education, Key Laboratory of Marine
Intelligent Equipment and System of Ministry of Education, Shanghai Engineering Research Center of Intelligent Control and Management, Shanghai
Jiao Tong University, Shanghai 200240, China.}%
\thanks{$^{2}$Dimanshen Technology Co., Ltd. specializes in 3D SLAM and robotic vision fusion technology, offering all-terrain intelligent robotic solutions for smart security and smart campus applications.}%
\thanks{$^{3}$Engineering Research Center of Intelligent Control for Underground Space, Ministry of Education, School
of Information and Control Engineering, Advanced Robotics Research
Center, China University of Mining and Technology, Xuzhou 221116,
China.}%
}

\begin{document}

\begin{CJK}{UTF8}{gbsn}
\maketitle
\thispagestyle{empty}
\pagestyle{empty}

\begin{abstract}

Dynamic scene reconstruction for autonomous driving enables vehicles to perceive and interpret complex scene changes more precisely. Dynamic Neural Radiance Fields (NeRFs) have recently shown promising capability in scene modeling. However, many existing methods rely heavily on accurate poses inputs and multi-sensor data, leading to increased system complexity. To address this, we propose FreeDriveRF, which reconstructs dynamic driving scenes using only sequential RGB images without requiring poses inputs. We innovatively decouple dynamic and static parts at the early sampling level using semantic supervision, mitigating image blurring and artifacts. To overcome the challenges posed by object motion and occlusion in monocular camera, we introduce a warped ray-guided dynamic object rendering consistency loss, utilizing optical flow to better constrain the dynamic modeling process. Additionally, we incorporate estimated dynamic flow to constrain the pose optimization process, improving the stability and accuracy of unbounded scene reconstruction. Extensive experiments conducted on the KITTI and Waymo datasets demonstrate the superior performance of our method in dynamic scene modeling for autonomous driving. Our implementation will be
available at https://github.com/IRMVLab/FreeDriveRF.

\end{abstract}

\section{INTRODUCTION}

\begin{figure}[t]
    \centering
    \includegraphics[width=0.5\textwidth]{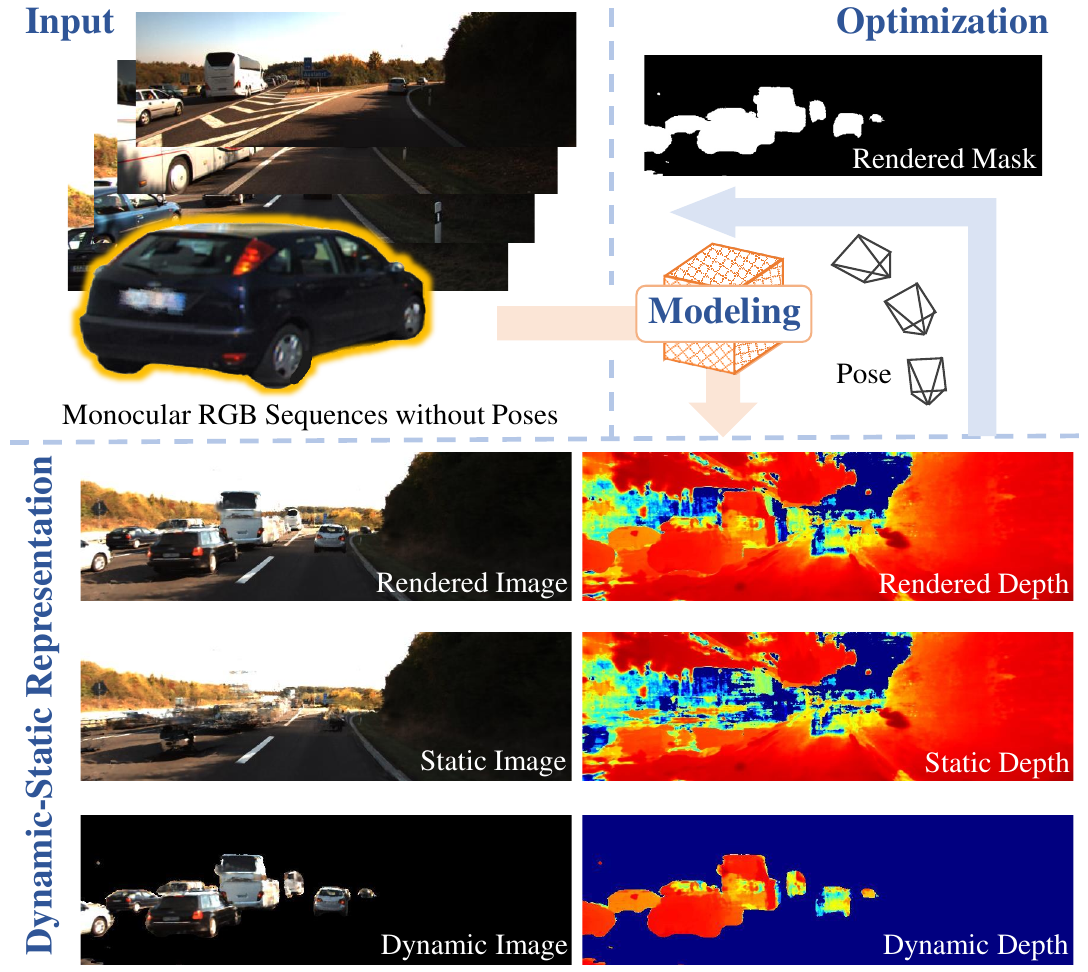}
    \caption{Our method reconstructs autonomous driving scenes from monocular RGB sequences without ground truth poses. During optimization, camera poses and rendered masks are updated, guiding dynamic modeling. At the bottom are the rendered RGB and depth maps for both dynamic and static components.}
    \label{fig:1-1}
    \vspace{-5mm} 
\end{figure}

\begin{figure*}[t]
    \centering
    \includegraphics[width=\textwidth]{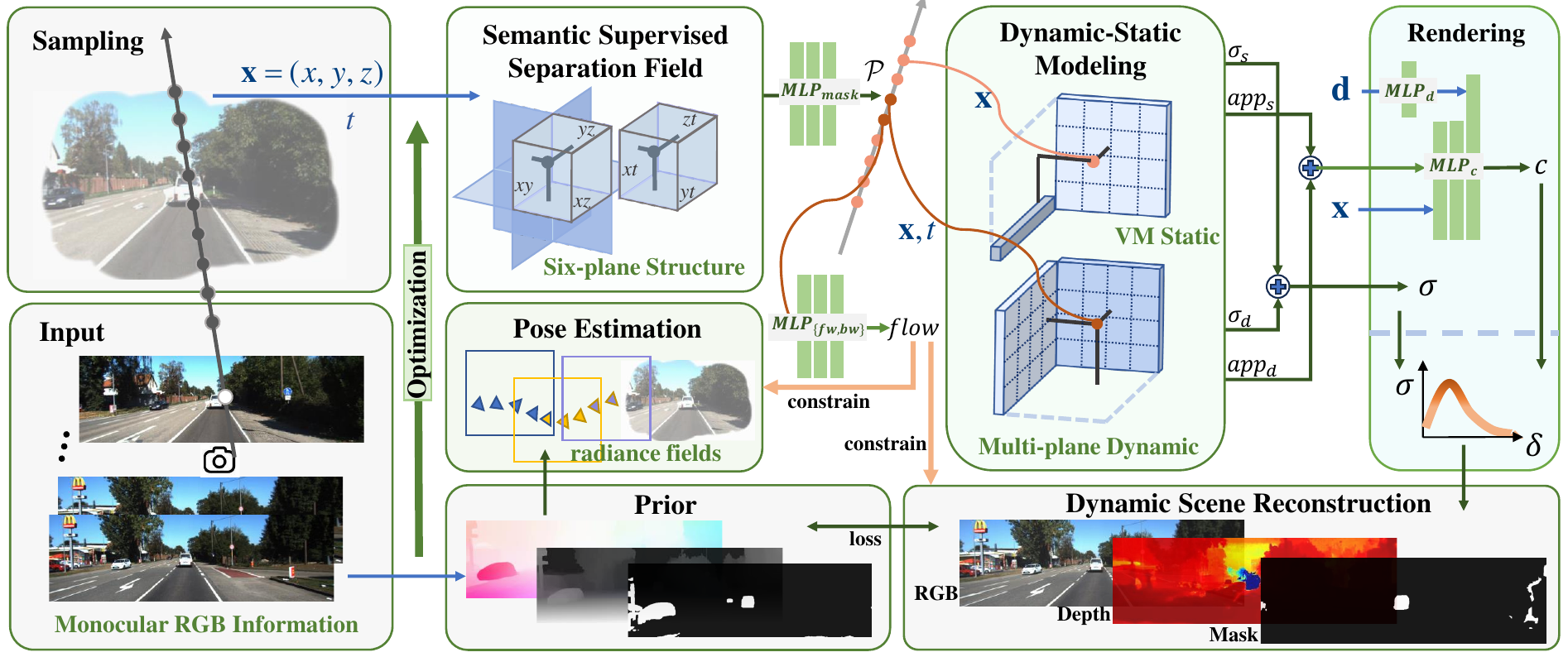}
    \caption{\textbf{FreeDriveRF Overview.} Our method processes monocular RGB sequences by first sampling rays and inputting points into a dynamic-static separation field, generating a probability $\mathcal{P}$ for each point under mask supervision to distinguish between static and dynamic points for separate modeling. Meanwhile, a scene flow field $\text{MLP}_{\{\text{fw}, \text{bw}\}}$ from dynamic part guides optimization and supervision process. The densities and appearance features are combined with the view direction to compute the final color. Volumetric rendering produces three maps supervised by ground truth and priors, with camera poses jointly optimizing with radiance fields.}
    \label{fig:method}
    \vspace{-3mm} 
\end{figure*}
Dynamic scene reconstruction supports applications in simulation and scenario replay for autonomous driving. Traditional methods, such as geometry-based Structure from Motion (SfM) \cite{schonberger2016structure} and deep learning-based Multi-View Stereo (MVS) \cite{yao2018mvsnet}, achieve dynamic scene reconstruction by detecting and removing dynamic objects. However, they fail to accurately modeling dynamic processes.

Recently, NeRF \cite{mildenhall2020nerf} has demonstrated impressive capability in dynamic scene modeling. However, it faces significant challenges in large-scale dynamic scenes. Previous dynamic-static methods struggle to decouple dynamic elements, leading to boundary artifacts and holes from linear fusion. Many dynamic reconstruction approaches also rely on multi-sensor inputs like cameras and LiDAR, complicating the system. Furthermore, some algorithms achieve pose estimation but fail to generalize to outdoor dynamic scenes \cite{pumarola2021d} or require strict camera motion constraints \cite{liu2023robust}.

To address these limitations, we propose FreeDriveRF, a novel dynamic NeRF reconstruction method that requires only monocular RGB image sequences as input for outdoor autonomous driving scenes without poses inputs. To solve the difficulty in separating dynamic and static elements, our approach decouples them at the sampling level using a mask-supervised semantic separation field. This process assigns dynamic and static points to independent models, effectively alleviating artifacts. To tackle the issues caused by moving objects and occlusions, we utilize optical flow between adjacent frames to track and align rays with object motion, ensuring rendering consistency. Furthermore, we incorporate dynamic scene flow into the joint optimization of camera poses and radiance fields, avoiding the information loss caused by naively masking dynamic objects and enhancing pose estimation accuracy. We evaluate our approach on KITTI and Waymo, demonstrating superior performance in both pose optimization and dynamic reconstruction tasks.

In summary, our contributions are as follows:
\begin{itemize}
\item We propose FreeDriveRF, a novel dynamic scene reconstruction algorithm that solely relies on monocular RGB image sequences as input without ground truth poses.
\item We introduce an innovative framework that decouples dynamic and static points at the sampling level, incorporating a warped ray-guided dynamic object consistency strategy to model dynamic elements more effectively.
\item We introduce dynamic object flow constraints in the joint optimization of camera poses and radiance fields, significantly enhancing reconstruction accuracy in dynamic scenes while improving pose optimization results.
\end{itemize}

\section{Related Work}

\subsection{Dynamic Scene Modeling}
Most dynamic NeRF methods use time as an additional input for scene reconstruction \cite{pumarola2021d,xian2021space,gao2021dynamic,park2021nerfies,tretschk2021non}. \text{D}$^2$\text{NeRF} \cite{wu2022d2nerf} learns a 3D representation of the scene from a monocular video, decoupling moving objects from a static background. HyperNeRF \cite{park2021hypernerf} handles topological changes by elevating NeRF to higher dimensions of space, resulting in more realistic renderings. The reconstruction of large-scale autonomous driving scenes has attracted the attention of many researchers \cite{turki2022mega,wu2023mars,lu2023urban,liu2023real,deng2023prosgnerf}. EmerNeRF \cite{yangemernerf} couples static, dynamic, and guided flow fields together to self-sustainably represent highly dynamic scenarios. SUDS \cite{turki2023suds} leverages unlabeled inputs to learn semantic awareness and scene flow, enabling it to perform multiple downstream tasks. HexPlane \cite{cao2023hexplane} employs six learned feature planes as a grid-based representation to explicitly encode spatial-temporal features, significantly accelerating training. With the rise of 3D Gaussian Splatting \cite{kerbl3Dgaussians,zhu20243d}, research in large-scale dynamic fields has emerged \cite{zhou2024drivinggaussian,huang2024textit}, but these approaches still depend on real poses or multiple sensors.

\subsection{Camera Pose Estimation }
NeRF relies on accurate camera poses from SfM or COLMAP \cite{schonberger2016pixelwise}, which struggles with large motion angles and blur, making pose-free NeRF a key research focus. NeRF\texttt{--} \cite{wang2021nerf} introduces direct intra-camera reference optimization for multi-view reconstruction, while Barf \cite{lin2021barf} efficiently corrects pose misalignment. However, both methods fail with large-scale video sequences. Nice-slam \cite{zhu2022nice} and Vox-Fusion \cite{yang2022vox} perform well in pose estimation but rely on RGB-D inputs and require precise depth. RoDynRF \cite{liu2023robust} optimizes poses in dynamic scenes from monocular sequences but is restricted by limited camera motion. LocalRF \cite{meuleman2023progressively} jointly estimates poses and radiance fields but is limited to static scenes. Several approaches leverage semantic information to enhance pose optimization \cite{zhu2024sni,zhu2024semgauss}. Our work integrates dynamic object flows into pose optimization, enabling effective reconstruction in large-scale dynamic scenes.

\section{Method}

Fig. \ref{fig:method} illustrates the overview of FreeDriveRF. Firstly, Sec.~\ref{sec:A} briefly introduces NeRF and our scene representation. Secondly, Sec.~\ref{sec:B} elaborates on how the proposed semantically supervised dynamic-static separation field enables the decoupling and modeling of complex scenes and introduces a more effective rendering method. Next, Sec.~\ref{sec:C} describes how the neural flow field and prior information constrain dynamic objects during pose estimation. Finally, Sec.~\ref{sec:D} explains the principle of object tracking-based spatiotemporal rendering consistency. Sec.~\ref{sec:E} derives the training loss.

\subsection{Preliminaries}
\label{sec:A}

NeRF synthesizes images by sampling 5D coordinate positions \(\mathbf{x} = (x, y, z)\) and viewing directions \(\mathbf{d} = (\theta, \phi)\) along rays. These are fed into an \( \text{MLP}_\Phi \) to produce color \(\mathbf{c}\) and density \(\sigma\), which are used for volumetric rendering:
\begin{equation}
\text{MLP}_\Phi : (\mathbf{x}, \mathbf{d}) \rightarrow (\mathbf{c}, \sigma).
\end{equation}
The pixel color is computed by integrating \(N\) sampled points along a ray \(\mathbf{r}\):
\begin{equation}
\label{eq:render}
\hat{\mathbf{C}}(\mathbf{r}) = \sum_{i=1}^{N} T_i (1 - \exp(-\sigma_i \delta_i)) \mathbf{c}_i,
\end{equation}
where \( T_i = \exp\left(-\sum_{j=1}^{i-1} \sigma_j \delta_j \right) \) represents the transmittance, and \( \delta_i \) is the distance between points.

For scene representation, We integrate VM decomposition \cite{chen2022tensorf} and multi-plane \cite{cao2023hexplane} structures to capture space-time information efficiently. We also non linearly map unbounded scenes into a cubic space (side length of 4) following \cite{meuleman2023progressively} and \cite{barron2022mip}, and remap timestamps \( t \) to \([-2, 2]\).

\subsection{Dynamic-Static Scene Decomposition and Reconstruction}
\label{sec:B}

We introduce a learnable 4D semantic supervision field to decouple dynamic and static sampling points along each ray, allowing independent modeling and reducing artifacts caused by mutual interference. By incorporating high-dimensional viewpoints and upsampling the grid, we capture richer temporal dynamics, enhancing reconstruction quality.


\noindent\textbf{Separation Field.}
Inspired by HexPlane \cite{cao2023hexplane}, for each sampling point \( \mathbf{P} = (x, y, z, t) \) with ray direction \( \mathbf{d} \), we obtain mask features from six planes. This separation field \( \mathbf{V}_m \) aggregates spatial and temporal information across time:
\begin{equation}
\mathbf{V}_m(x, y, z, t) = \sum_{((i,j), (k,l))} \sum_{r=1}^{R_m} \mathbf{M}_r^{i,j} \circ \mathbf{M}_r^{k,l},
\end{equation}
where \( ((i,j), (k,l)) \) represents pairs of coordinate axes, i.e., \( (XY, ZT), (XZ, YT), (YZ, XT) \), and each \( \mathbf{M} \) is a set of learned feature planes. \( \circ \) represents the outer product and \( R_m \) denotes the number of planes. After separation, we apply \(\text{MLP}_{\text{mask}}\) to compute the dynamic probability \(\mathcal{P}\) for each point. A learnable threshold \(\tau\), constrained by \( L_2 \) regularization, is used to classify points:
\begin{equation}
\mathcal{L}_{\tau} = \left( \tau - 0.5 \right)^2,
\end{equation}
where points with \(\mathcal{P} > \tau\) are dynamic and those with \(\mathcal{P} \leq \tau\) are static, assigning them to their respective fields.

\noindent\textbf{Static Field.}
For static points, we only need to utilize the position to model the static field by vector-matrix products:
\begin{equation}
\mathbf{V}_{\{ \sigma, \mathbf{c} \}}^s (x, y, z) = \sum_{(i, (j,k))} \sum_{r=1}^{R_{\{ \sigma, \mathbf{c} \}}} \mathbf{v}_r^i \circ \mathbf{M}_r^{j,k},
\end{equation}
where \(\mathbf{V}_{\{ \sigma, \mathbf{c} \}}^s \) represent the static density and appearance field respectively. The axes \( (i, (j,k)) \) are \( (X, YZ), (Y, XZ), (Z, XY) \), and \( \mathbf{v} \) is a learnable vector.

\noindent\textbf{Dynamic Field.}
Due to the need to recover moving elements across the time dimension, the dynamic density field  \( \mathbf{V}_{\sigma}^d \) and appearance field \( \mathbf{V}_{\mathbf{c}}^d \) are modeled using the six-plane structure similar to the dynamic-static separation field. A dynamic field is composed of $6R_\sigma + 6R_\mathbf{c}$ planes.

    
    


\begin{figure}[t]
    \centering
    \includegraphics[width=0.5\textwidth]{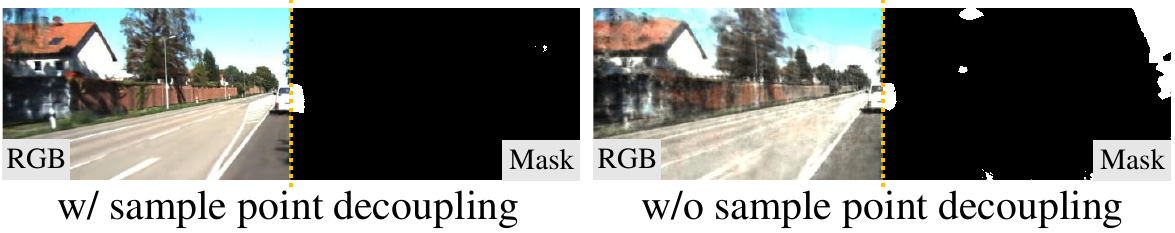}
    \setlength{\abovecaptionskip}{-8pt} 
    \setlength{\belowcaptionskip}{8pt} 
    \caption{\textbf{Visual comparison of rendered RGB and masks with or without the proposed sampling level dynamic-static decoupling}.}
    \label{fig:decoupling}
    \vspace{-2mm} 
\end{figure}

\begin{figure}[t]
    \centering
    \includegraphics[width=0.5\textwidth]{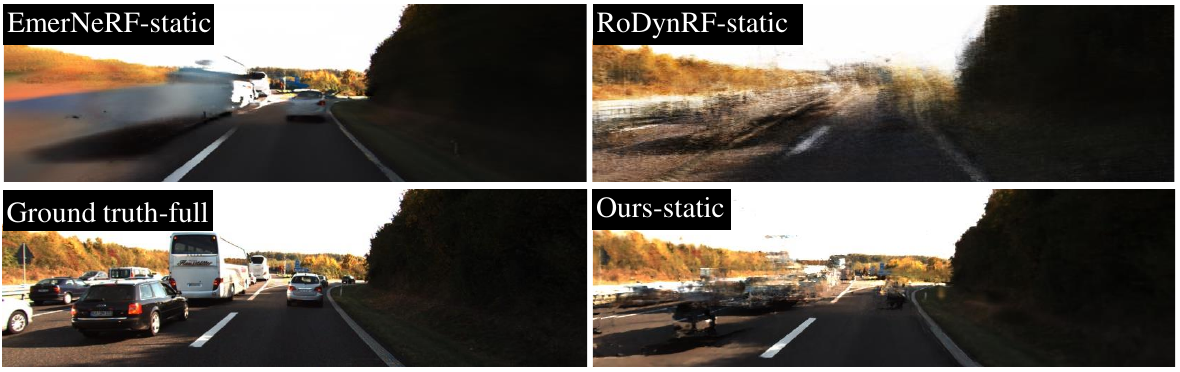}
    \setlength{\abovecaptionskip}{-8pt} 
    \setlength{\belowcaptionskip}{8pt} 
    \caption{\textbf{Static background reconstruction.} Our sampling point level dynamic-static decoupling reconstructs occluded static regions more effectively and produces fewer artifacts compared to others.  }
    \label{fig:static_modeling}
    \vspace{-6mm} 
\end{figure}

\noindent\textbf{Ray Aggregation and Rendering.}
After modeling, we obtain static and dynamic density \( \sigma_s \), \( \sigma_d \), and appearance information \( app_s \), \( app_d \) through trilinear and bilinear interpolation on the multi-resolution feature grids \(\mathbf{V}_{\{ \sigma, \mathbf{c} \}}^s\) and \(\mathbf{V}_{\{ \sigma, \mathbf{c} \}}^d\). To ensure that all points can represent respective features during rendering, we aggregate the density and appearance features of each dynamic and static point directly along the ray:
\begin{equation}
\sigma = \text{concat}(\sigma_s, \sigma_d), app = \text{concat}(app_s, app_d),
\end{equation}
\begin{equation}
\mathbf{c} = \text{MLP}_{\text{c}}(\text{PE}(\mathbf{x}), \text{MLP}_{\text{d}}(\text{PE}(\mathbf{d})), app),
\end{equation}
where \(\text{PE}(\cdot)\) represents the position encoding. The final features \( \sigma \) and \( \mathbf{c} \) are rendered to obtain the pixel colors \(\hat{\mathbf{C}}(\mathbf{r})\) through \eqref{eq:render}. Particularly, before passing through the final \(\text{MLP}_{\text{c}}\), the viewing direction \(\mathbf{d}\) is first processed by an \(\text{MLP}_{\text{d}}\) to better capture the effect of view changes on scene appearance, enabling more detailed modeling of dynamic elements from different viewpoints.

\noindent\textbf{Upsampling.} Inspired by TensoRF \cite{chen2022tensorf}, we adopt a coarse-to-fine grid optimization strategy during training, which helps the network capture small-scale deformations and distinguish dynamic from static elements, thereby improving object scale estimation and motion modeling accuracy.

As shown in Fig. \ref{fig:decoupling}, our method enhances the ability to separate static and dynamic semantics, improving rendering quality while significantly reducing dynamic artifacts.

\subsection{Pose Estimation with Dynamic Objects}
\label{sec:C}

\setlength{\tabcolsep}{2mm}
\begin{table}[t]
    \begin{center}
        \caption{\textbf{Quantitative comparison with other NeRF-based pose optimization algorithms on KITTI.} Both our scene reconstruction and pose optimization outperform others.}
        \label{tab:pose}
        \begin{tabular}{l|cc|cc}
            \toprule
            Method & PSNR\(\uparrow\) & SSIM\(\uparrow\) & ATE(m)\(\downarrow\) & RTE(m)\(\downarrow\)\\
            \midrule
            LocalRF \cite{meuleman2023progressively} & 22.68 & 0.665 & 1.9548 & 0.1696\\
            RoDynRF \cite{liu2023robust} & 20.32 & 0.587 & 13.2256 & 1.1163\\
            Ours w/ flow constraint & \textbf{24.21} & \textbf{0.699} & \textbf{1.0577} & \textbf{0.1460} \\
            \midrule
            Ours w/o flow constraint & 23.67 & 0.597 & 2.2546 & 0.1946\\
            \bottomrule
        \end{tabular}
    \end{center}
    \vspace{-2mm} 
\end{table}

\begin{figure}[t]
    \centering
    \includegraphics[width=0.5\textwidth]{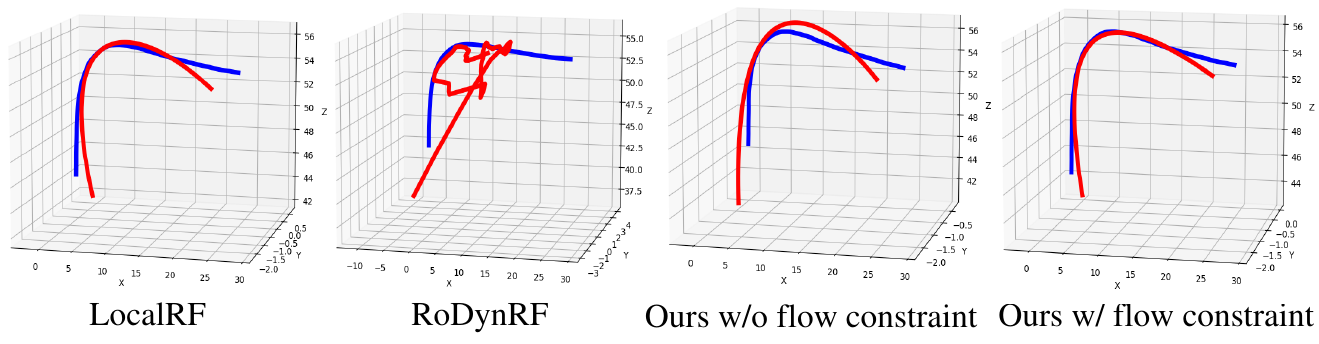}
    \caption{\textbf{Comparison of pose trajectory on sequence 03 of KITTI.} We present our qualitative results compared with \cite{liu2023robust}, \cite{meuleman2023progressively}, and without optical flow constraint for pose optimization.}
    \label{fig:psoe}
    \vspace{-4mm} 
\end{figure}

In large-scale dynamic environments, moving objects complicate pose estimation. LocalRF \cite{meuleman2023progressively} refines poses and local fields but is limited to static scenes. We extend it to dynamic settings by introducing a scene flow field and leveraging 2D flow and monocular depth priors for supervision.

\noindent\textbf{Depth Loss.} 
We use DPT \cite{ranftl2021vision} to compute the inverse depth \( \mathbf{D} \), represented as a grayscale image. The predicted depth \( \hat{\mathbf{D}}(\mathbf{r}) \) defines a loss that accounts for scale and translation variations. The corresponding inverse depth \(\hat{\mathbf{D}}(\mathbf{r})\) is:
\begin{equation}
\quad \hat{\mathbf{D}}(\mathbf{r}) = \sum_{i=1}^{N} T_i (1 - \exp(-\sigma_i \delta_i)) d_i, \hat{\mathbf{D}}_{\text{inv}} = \frac{1}{\hat{\mathbf{D}} + \epsilon},
\end{equation}
where \( d_i \) is the depth of the \( i \)-th sampled point of \( N \), and \(\epsilon\) is a small constant. To ensure scale invariance, we apply median normalization \( \mathcal{N}(\cdot) \) to normalizes inverse depth. The depth loss is \(\mathcal{L}_{\text{d}} = \left| \mathcal{N}( \hat{\mathbf{D}}_{\text{inv}}) - \mathcal{N}(\mathbf{D}) \right|^2.\)

\noindent\textbf{Object Flow Constraint.}
\label{sec:Flow}
We model the scene flow field \(\text{MLP}_{\text{\{fw, bw\}}}\) for the dynamic points, predicting 3D motion flow based on time, position, and encoded features:
\begin{equation}
fl_{\{ f, b \}} = \text{MLP}_{\text{\{fw, bw\}}}(\mathbf{x}, \text{PE}(\mathbf{x}), t, \text{PE}(t)).
\end{equation}
Then we obtain pseudo ground truth 2D flow \( \mathbf{F}_{q \rightarrow q+1} \) for \( q \in [1..Q-1] \) between frames \cite{teed2020raft}. For the expected flow, the forward \( fl_f \) is subtracted from the transformed 3D points:
\begin{equation}
\hat{\mathbf{F}}_{q \rightarrow q+1} = p - \Pi \left( [R|t]_{q \rightarrow q+1} \cdot \Pi^{-1}(p, \hat{\mathbf{D}}) - fl_f\right),
\end{equation}
where  \( p \) represents pixel coordinates. \( \Pi \) represents the projection from 3D points to image space and \( \Pi^{-1} \) denotes its inverse, which reconstructs 3D points using depth. \( [R|t]_{q \rightarrow q+1} \) is the relative camera pose from the \( q \)-th to the \( q+1 \)-th frame. The forward and backward flow loss, \( \mathcal{L}_{\text{flow}}^f \) and \( \mathcal{L}_{\text{flow}}^f \) are the L1 norm between 2D predicted flow and ground truth. We compute the L1 loss \( \mathcal{L}_{\text{flow}}^{fb} \) as the sum of the forward 3D optical flow and the next frame's backward optical flow.

By incorporating dynamic flow into pose optimization, our approach surpasses methods that mask out dynamic objects and solely rely on the static background, resulting in more accurate pose recovery. We follow the setup of \cite{meuleman2023progressively} for adding local radiance fields and repeat until the entire trajectory is covered, producing a complete reconstruction.

\subsection{Dynamic Objects Spatiotemporal Rendering Consistency}
\label{sec:D}

\begin{figure}[t]
    \centering
    \includegraphics[width=0.5\textwidth]{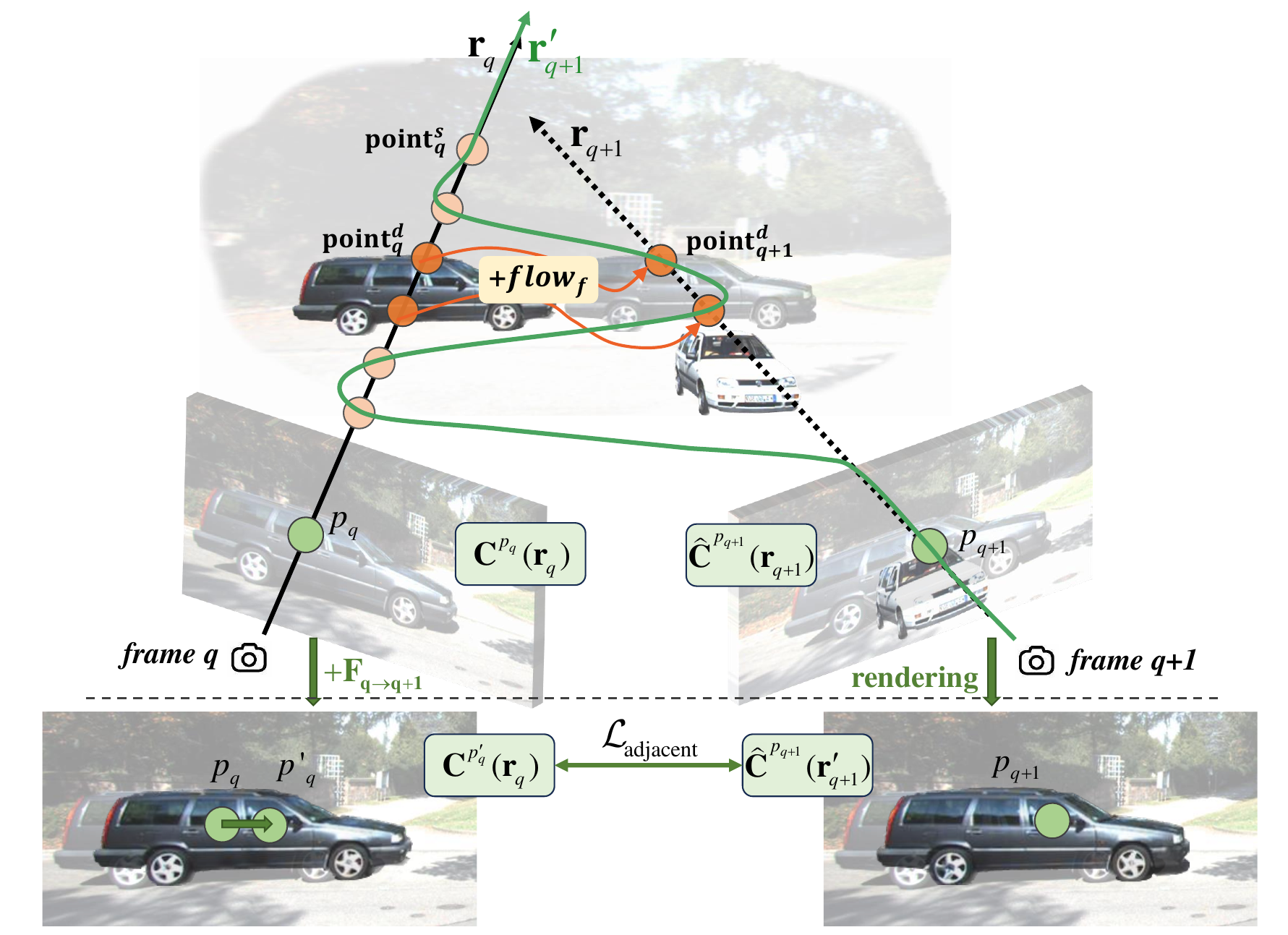}
    \caption{\textbf{Warped ray-guided dynamic object rendering consistency.} Dynamic points are shifted from frame \( q \) to \( q+1 \) using the predicted 3D flow \( fl_f \), generating a warped ray \( \mathbf{r'}_{q+1} \). This ray bypasses the white car foreground and passes through the static background in frame \( q \) and the dynamic black car in frame \( q+1 \). The final pixel color through \( \mathbf{r'}_{q+1} \) is \( \hat{\mathbf{C}}^{p_{q+1}}(\mathbf{r'}_{q+1}) \), instead of \( \hat{\mathbf{C}}^{p_{q+1}}(\mathbf{r}_{q+1}) \). The 2D flow \( \mathbf{F}_{q \rightarrow q+1} \) aligns dynamic pixels from \( p_q\) to \( p_q'\) in frame \( q \), constructing a loss to constrain dynamic modeling.}
    \label{fig:flow}
    \vspace{-5mm} 
\end{figure}

\setlength{\tabcolsep}{2mm}
\begin{table*}[t]
    \begin{center}
        \caption{\textbf{Quantitative results of novel view synthesis on KITTI sequences.}}
        \label{tab:kitti}
        \begin{tabular}{l|c|cccccc|c}
            \toprule
            PSNR\(\uparrow\) / LPIPS\(\downarrow\) & input poses & 03 & 04 & 05 & 09 & 18 & 20 & Average \\
            \midrule
            \text{D}$^2$\text{NeRF} \cite{wu2022d2nerf} & Yes & 18.02 / 0.382 & 20.99 / 0.366 & 22.56 / 0.418 & 21.87 / 0.499 & 19.28 / 0.396 & 22.73 / 0.218 & 20.91 / 0.380\\
            RoDynRF \cite{liu2023robust} & No & 18.82 / 0.524 & 21.57 / 0.504 & 20.47 / 0.523 & 22.34 / 0.424 & 18.54 / 0.386 & 23.25 / 0.385 & 20.83 / 0.458\\
            EmerNeRF \cite{yangemernerf} & Yes & 23.82 / \underline{0.343} & \underline{24.82} / 0.343 & \underline{23.35} / 0.421 & \underline{24.41} / 0.394 & \underline{21.76} / 0.507 & \textbf{29.34} / \underline{0.159} & \underline{24.58} / 0.361\\
            LocalRF \cite{meuleman2023progressively} & No & 22.44 / \textbf{0.156} & 22.80 / 0.363 & 20.73 / 0.411 & 21.43 / \underline{0.354} & 21.35 / 0.371 & 23.74 / 0.253 & 22.08 / 0.318\\
            Ours w/ pose & No & \textbf{24.04} / \underline{0.244} & \textbf{25.04} / \textbf{0.227} & \textbf{23.43} / \textbf{0.314} & \textbf{24.66} / \textbf{0.342} & \textbf{22.93} / \textbf{0.290} & \underline{28.57} / \textbf{0.152} & \textbf{24.78} / \textbf{0.262}\\
            \midrule
            Ours w/o pose & Yes & 22.12 / 0.268 & 22.53 / \underline{0.284} & 21.59 / \underline{0.378} & 20.14 / 0.366 & 21.02 / \underline{0.303} & 26.83 / 0.165 & 22.37 / \underline{0.294}\\
            \bottomrule
        \end{tabular}
    \end{center}
    \vspace{-5mm} 
\end{table*}

We propose a warped ray-guided consistency strategy to improve dynamic object modeling by enforcing temporal consistency. Instead of addressing occlusions directly, our method uses warped rays to bypass them, creating a custom loss for frame-to-frame consistency in Fig. \ref{fig:flow}, leveraging 3D scene flow for rendering and 2D optical flow for supervision.

\noindent\textbf{3D Flow for Dynamic Guiding.} Specifically, after obtaining the forward \( fl_f \) from Sec.~\ref{sec:Flow}, we transform the 3D dynamic points recovered from the rendered depth map in frame \( q \) to frame \( q + 1 \) as \( \text{point}_{q+1}^d = \text{point}_q^d + fl_f \), while keeping the static unchanged. Through this warping,  we can obtain a new ray \( \mathbf{r'}_{q+1}\) with direction \( \mathbf{r'}_d^{q+1} = \text{point}_{q+1} - \mathbf{r}_o^{q+1} \) from the camera origin at frame \( q+1 \) to \( \text{point}_{q+1} \).

\noindent\textbf{Warped Ray Rendering.} The core is that the trajectory of warped rays continuously follows the dynamic objects present in the current frame, so \( \mathbf{r'}_{q+1}\) passes through static elements from frame \( q \) and dynamic objects from frame \( q+1 \) that have appeared in frame \( q \). Using the standard rendering process along \( \mathbf{r'}_{q+1}\), we compute the pixel color \( \hat{\mathbf{C}}^{p_{q+1}}(\mathbf{r'}_{q+1}) \). This bypasses the foreground that appears to block the background when viewing from the initial ray \( \mathbf{r}_{q+1}\). 

\noindent\textbf{2D Flow for Pixel Alignment.} However, since the rendering is done from the perspective of frame \( q+1 \), it cannot be directly compared to the ground truth of the current frame \(\mathbf{C}^{p_q}(\mathbf{r}_{q})\). To align the pixel, we use the 2D flow \( \mathbf{F}_{q \rightarrow q+1} \) to map the coordinate \( p_q \) to \( p_q' = p_q + \mathbf{F}_{q \rightarrow q+1} \) in frame \( q \). Thus, the final loss can be:
\begin{equation}
\mathcal{L}_{\text{adjacent}} = \sum_{r} \left\| \hat{\mathbf{C}}^{p_{q+1}}(\mathbf{r'}_{q+1}) - \mathbf{C}^{p'_q}(\mathbf{r}_{q}) \right\|_2^2.
\end{equation}
This ensures dynamic motion invariance across time, enhancing reconstruction accuracy, even at dynamic boundaries.

\begin{figure}[t]
    \centering
    \includegraphics[width=0.5\textwidth]{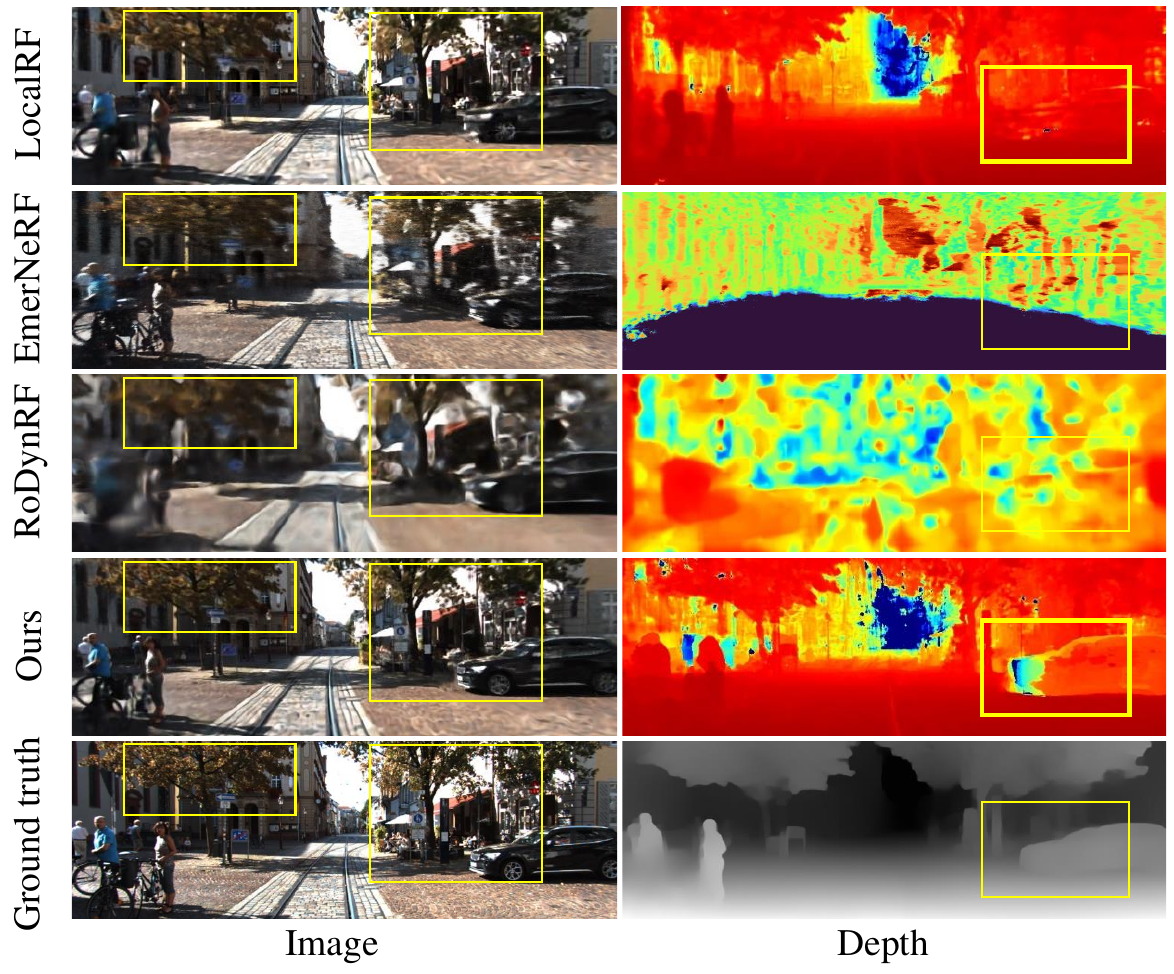}
    \caption{\textbf{Qualitative comparison results on KITTI.} The left column shows the rendered images, and the right column displays the depth.}
    \label{fig:kitti-depth}
    \vspace{-4mm} 
\end{figure}

\subsection{Total Loss}

\label{sec:E}
During the training process, we employ the L2 loss  \(\mathcal{L}_{\text{rgb}} \) for the predicted pixel colors， \(\mathcal{L}_{\tau} \) for the learnable threshold, and \(\mathcal{L}_{\text{d}} \) for the normalized depth, the L1 loss for optical flow \(\mathcal{L}_{\text{flow}} = \mathcal{L}_{\text{flow}}^f + \mathcal{L}_{\text{flow}}^b \) + \(\mathcal{L}_{\text{flow}}^{fb} \), and dynamic rendering consistency loss \(\mathcal{L}_{\text{adjacent}} \). We also combine Mask R-CNN \cite{he2017mask} and Sampson distance to obtain the pseudo ground truth motion mask and render the semantic maps of the prediction:
\begin{equation}
\hat{\text{Mask}}(\mathbf{r}) = \sum_{i=1}^{N} T_i (1 - \exp(-\sigma_i \delta_i)) \mathcal{P}_i,
\end{equation}
where \( \mathcal{P}_i \) is the estimated dynamic probability mask. Then binary cross-entropy loss \( \mathcal{L}_{\text{mask}} \) is adopted to supervise the dynamic-static separation field. Finally, we minimize:
\begin{equation}
\label{eq:loss}
\begin{split}
\mathcal{L} = & \ \lambda_1 \mathcal{L}_{\text{rgb}} + \lambda_2 \mathcal{L}_{\text{d}} + \lambda_3 \mathcal{L}_{\text{flow}} \\
& + \lambda_4 \mathcal{L}_{\text{mask}} + \lambda_5 \mathcal{L}_{\text{adjacent}} + \lambda_6 \mathcal{L}_{\tau},
\end{split}
\end{equation}
where \( \lambda_1 \) to \( \lambda_6 \) are hyperparameters.

\begin{figure*}[t]
    \centering
    \begin{subfigure}[t]{\textwidth}
        \centering
        \includegraphics[width=\textwidth]{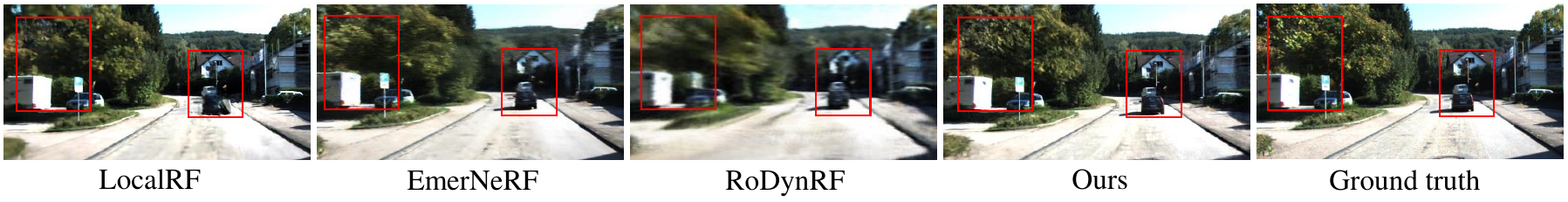}
        \captionsetup{skip=-0.005cm} 
        \caption{}
        \label{fig:kitti}
    \end{subfigure}
    
    \vspace{0.1cm} 
    
    \begin{subfigure}[t]{\textwidth}
        \centering
        \includegraphics[width=\textwidth]{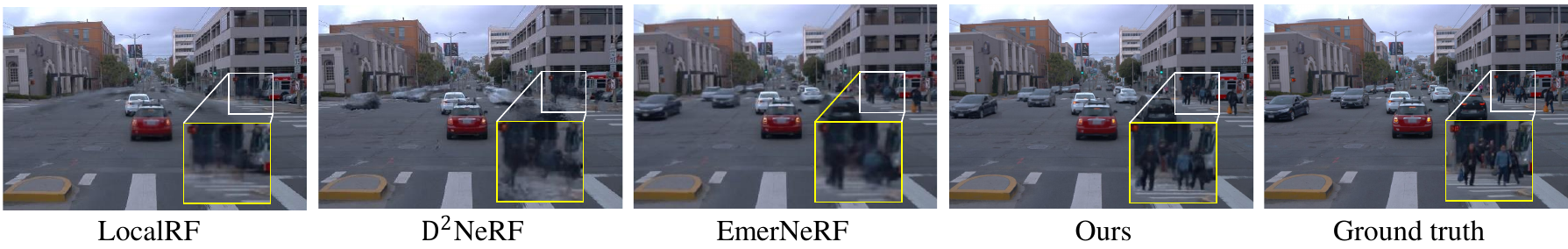}
        \captionsetup{skip=-0.02cm}
        \caption{}
        \label{fig:waymo}
    \end{subfigure}

    \vskip -0.2cm 
    \caption{\textbf{Comparison of novel view synthesis results.} We present the detailed portions of synthesized images from novel views on the (a) KITTI and (b) NOTR datasets. Our method more effectively reconstructs distant dynamic vehicles and pedestrians, while also producing superior ground textures and vegetation details.}
    \label{fig:combined_example}
    \vspace{-3mm} 
\end{figure*}


\section{Experiments}
\subsection{Experiment Setup}
\noindent\textbf{Implementation Details.}
The optimization process begins with a dynamic-static radiance field model integrated with pose optimization. We use the Adam optimizer for all parameters, training for one day on an RTX 8000 GPU. Initial regularization weights for flow, depth, and mask are set to 1, 0.5, and 0.5, respectively. For the upsampling of spatial grids, we uniformly start at \( 64^3 \). The loss weights \( \lambda_1 \) to \( \lambda_6 \) in \eqref{eq:loss} are set to 0.7, 1, 1, 1, 0.5, and 0.5.

\noindent\textbf{Datasets.}
We use KITTI \cite{geiger2012we} and a subset of the Waymo Open dataset \cite{sun2020scalability}, NOTR \cite{yangemernerf} for evaluation. From KITTI, seven highway and urban driving sequences are sampled. And dynamic-32 and static-32 from NOTR represent complex urban traffic scenarios. Consecutive frames from the left color camera in KITTI and the front-view camera in NOTR are used to simulate monocular sequences.

\noindent\textbf{Metrics.}
We assess scene reconstruction and novel view synthesis using Peak Signal-to-Noise Ratio (PSNR), Structural Similarity Index (SSIM), and Learned Perceptual Image Patch Similarity (LPIPS). Pose optimization performance is evaluated with Absolute Trajectory Error (ATE) and Relative Translation Error (RTE).

\setlength{\tabcolsep}{5mm}
\begin{table}[t]
    \begin{center}
        \caption{\textbf{Quantitative results of novel view synthesis on NOTR sequences.}}
        \label{tab:notr}
        \begin{tabular}{l|ccc}
            \toprule
            PSNR\(\uparrow\) / LPIPS\(\downarrow\) & Dynamic-32 & Static-32\\
            \midrule
            HyperNeRF \cite{park2021hypernerf} & 24.48 / \underline{0.260} & 25.48 / 0.263\\
            RoDynRF \cite{liu2023robust} & 25.17 / 0.343 & 25.57 / 0.310 \\
            EmerNeRF \cite{yangemernerf} &  \underline{26.33} / 0.296 & \textbf{27.04} / \underline{0.181}\\
            LocalRF \cite{meuleman2023progressively} & 26.18 / 0.288 & 24.25 / 0.210\\
            Ours & \textbf{26.67} / \textbf{0.242} & \underline{26.91} / \textbf{0.178}\\
            \bottomrule
        \end{tabular}
    \end{center}
    \vspace{-4mm} 
\end{table}

\subsection{Poses Estimation}
We evaluate trajectory error and scene reconstruction results for pose-optimizing algorithms \cite{meuleman2023progressively} and \cite{liu2023robust} that do not require ground truth poses input on the KITTI dataset sequences. For scene reconstruction, we use all frames for training. Quantitative results are provided in Tab. \ref{tab:pose}, showing that our method outperforms other NeRF-based approaches capable of pose optimization both in terms of pose estimation and scene reconstruction in large-scale dynamic environments. Fig. \ref{fig:psoe} shows a visual comparison of a trajectory in one KITTI sequence. It is evident that \cite{liu2023robust}, which performs global optimization over the entire sequence, encounters significant pose errors during fast camera turns. In contrast, our approach optimizes newly added poses and radiance fields incrementally while employing the dynamic object motion flow supervision, preventing incorrect environmental data from influencing the pose optimization process.


\subsection{Dynamic Novel View Synthesis}
\noindent\textbf{Quantitative evaluation.}
To evaluate the rendering performance, we first compare our method with existing outdoor scenes or dynamic reconstruction baselines on KITTI \cite{geiger2012we} sequences for novel view synthesis. Our method, along with \cite{liu2023robust} and \cite{meuleman2023progressively}, does not require poses inputs. It's important to note that since sequences 18 and 20 lack pose information, we first recover the poses using \cite{schonberger2016pixelwise} for other methods. To maintain monocular images as input, we disable LiDAR-related inputs in \cite{yangemernerf} with DINO feature retained. In large-scale dynamic scenes, the quantitative results in Tab. \ref{tab:kitti} show that our method outperforms other dynamic novel view synthesis algorithms on most KITTI sequences, with the best overall average performance. The quantitative results on NOTR presented in Tab. \ref{tab:notr} show that our method performs on par with existing methods.

\setlength{\tabcolsep}{2mm}
\begin{table}[t]
    \begin{center}
        \caption{\textbf{Ablation studies for dynamic scene reconstruction.}}
        \label{tab:reconstruction}
        \begin{tabular}{l|ccc}
            \toprule
             & PSNR\(\uparrow\) & SSIM\(\uparrow\) & LPIPS\(\downarrow\)\\
            \midrule
            w/o sampling point decoupling & 20.12 & 0.611 & 0.464 \\
            w/o rendering consistency & 23.48 & 0.646 & 0.342 \\
            w/o upsampling & 19.13 & 0.563 & 0.477 \\
            w/o learnable \(\tau\) & 23.14 & 0.678 & 0.256 \\
            Ours & \textbf{25.04} & \textbf{0.727} & \textbf{0.227}\\
            \bottomrule
        \end{tabular}
    \end{center}
    \vspace{-6mm} 
\end{table}

\noindent\textbf{Qualitative Evaluation.}
To provide an intuitive analysis, we present the results of novel view synthesis for RGB and depth maps on the KITTI dataset in Fig. \ref{fig:kitti-depth}. Under the same configuration as in the quantitative evaluation, our method generates images with fewer artifacts and depth maps with more accurate object descriptions. EmerNeRF \cite{yangemernerf} struggles with depth estimation and detail recovery due to the lack of ground truth point clouds. More detailed comparisons on KITTI can be seen in Fig. \ref{fig:kitti}. The qualitative results on NOTR are shown in Fig. \ref{fig:waymo}. It is clear that HyperNeRF \cite{park2021hypernerf} and \text{D}$^2$\text{NeRF} \cite{wu2022d2nerf} almost fail to capture dynamic vehicle details, while our method significantly outperforms others in reconstructing distant pedestrians and vegetation textures.

Moreover, as shown in Fig. \ref{fig:1-1}, our method can effectively render dynamic and static scenes after training by decoupling these elements from the source. This enables clearer and more accurate dynamic object reconstructions, even in challenging conditions. And Fig. \ref{fig:static_modeling} shows that our method successfully reconstructs most of the static background occluded by dynamic objects compared to \cite{yangemernerf} and \cite{liu2023robust}.


\subsection{Ablation Study}

\setlength{\tabcolsep}{2.5mm}
\begin{table}[t]
    \begin{center}
        \caption{\textbf{Ablation studies for pose estimation.}}
        \label{tab:pose estimation}
        \begin{tabular}{l|ccc}
            \toprule
             & PSNR\(\uparrow\) & SSIM\(\uparrow\) & LPIPS\(\downarrow\)\\
            \midrule
            w/o dynamic flow constraint & 23.43 & 0.618 & 0.371\\
            w/o \(\text{MLP}_{\text{d}}\) & 23.37 & 0.483 & 0.366 \\
            Ours & \textbf{25.04} & \textbf{0.727} & \textbf{0.227}\\
            \bottomrule
        \end{tabular}
    \end{center}
    \vspace{-6mm} 
\end{table}

\noindent\textbf{Static-Dynamic Decoupling and Modeling.}
We propose a method for decoupling dynamic and static elements at the sampling point level using a separation field. Unlike performing linear blending within a unified model, our approach delivers superior qualitative and quantitative results on sequence 04 of KITTI in Fig. \ref{fig:decoupling} and Tab. \ref{tab:reconstruction}. Our decoupling technique more effectively separates dynamic and static components, significantly enhancing the reconstruction of static backgrounds. Furthermore, we assess the impact of adjacent frame rendering consistency loss, grid upsampling, and learnable \(\tau\) of dynamic-static separation. Grid upsampling significantly impacts dynamic scene reconstruction, especially in complex scenes with high-detail areas. Tab. \ref{tab:reconstruction} demonstrates that each component contributes to the decomposition and reconstruction of dynamic and static scenes, with the complete system performing the best.

\noindent\textbf{Pose Optimization.}
For dynamic scenes, Tab. \ref{tab:pose estimation} shows that using our estimated flow to constrain the pose improves reconstruction accuracy. The trajectory visualization in Fig. \ref{fig:psoe} and the quantitative comparison of trajectory error in Tab. \ref{tab:pose} also demonstrated that our method achieves the highest trajectory accuracy with flow constraint. Incorporating an extra view \(\text{MLP}_{\text{d}}\) before obtaining color features can also capture more inter-frame continuity, enhancing inter-frame continuity and aiding pose optimization.




\section{LIMITATIONS AND CONCLUSIONS}

Our method's reliance on optical flow and dynamic masks may lead to pose inaccuracies and artifacts, especially with fast-moving objects. It also assumes temporally consistent frames, and regions observed for short durations may suffer from reconstruction ambiguity.

We present a novel approach for dynamic scene reconstruction in autonomous driving using monocular sequences without poses. Our method introduces a sampling point-level dynamic-static decoupling mechanism to model dynamic and static elements, which reduces artifacts and occlusions. We also propose a warped ray-guided rendering strategy to supervise dynamic object modeling. By integrating optical flow constraints, we improve the accuracy of pose estimation and scene reconstruction. Experiments on KITTI and Waymo datasets show that our approach excels in dynamic scene modeling and pose optimization.




\bibliographystyle{IEEEtran}  
\bibliography{IEEEexample}

\end{CJK}
\end{document}